\documentclass[fleqn,12pt,nolineno]{wlscirep}
\usepackage{graphicx}
\usepackage{hyperref}
\usepackage[utf8]{inputenc}
\usepackage[T1]{fontenc}
\usepackage{subcaption}
\usepackage{booktabs}
\usepackage{tabularx}
\usepackage{array}
\usepackage{makecell}
\usepackage{float}

\usepackage{booktabs} 

\title{MSGL-Transformer: A Multi-Scale Global-Local Transformer for Rodent Social Behavior Recognition}

\author[1]{Muhammad Imran Sharif}
\author[1*]{Doina Caragea}

\affil[1]{Department of Computer Science, Kansas State University, Manhattan, KS, 66506, USA.}
\affil[ ]{*e-mail: dcaragea@ksu.edu}

\begin{abstract}

Recognition of rodent behavior is  important for understanding neural and behavioral mechanisms. Traditional manual scoring of behavior is time-consuming and prone to human error. Deep learning has helped in automating behavioral analysis; however, most approaches struggle to simultaneously capture both short, subtle motions and long behavioral patterns. We propose MSGL-Transformer, a Multi-Scale Global-Local Transformer for recognizing rodent social behaviors from pose-based temporal sequences. The model employs a lightweight transformer encoder with multi-scale attention, enabling it to capture motion dynamics across different temporal scales. Unlike existing skeleton-based transformers that rely on predefined spatial graphs or fixed temporal windows, the proposed architecture integrates parallel short-range, medium-range, and global attention branches to explicitly capture behavior dynamics at multiple temporal scales. In addition, we introduce a Behavior-Aware Modulation (BAM) block,  inspired by the channel reweighting mechanism of SE-Networks. This block modulates temporal embeddings to emphasize behavior-relevant features prior to the attention operation. We evaluate the model on two publicly available datasets: RatSI, which has 5 behavior classes with 12D pose inputs corresponding to rats, and CalMS21, which contains 4 behavior classes with 28D pose inputs  corresponding to mice. On RatSI, the MSGL-Transformer achieves a mean accuracy of 75.4\% and an F1-score of 0.745  across nine cross-validation splits, outperforming baseline models including TCN, LSTM, and Bi-LSTM. On CalMS21, the model achieves an accuracy of 87.1\%  and an F1-score of 0.8745, representing a +10.7\% improvement in average per-class accuracy compared to HSTWFormer.  The model also outperforms other keypoint-based methods including ST-GCN, MS-G3D, CTR-GCN, and STGAT. Notably, the same architecture is used for both datasets, with only the input dimensionality and number of classes adjusted. This demonstrates that the model generalizes well across different species and pose configurations.

\keywords{Rodent behavior recognition, Multi-scale attention, 
Transformer, Behavior-aware modulation, Deep learning, 
Cross-dataset generalization}

\end{abstract}

\begin{document}

\raggedbottom
\maketitle
\thispagestyle{empty}

\section{Introduction}

Animals are widely utilized as experimental models in neuroscience research~\cite{romanova2018animal}. Their behavioral responses provide useful information regarding social interaction, response to environmental stimuli, and decision-making patterns~\cite{zakowski2020animal}. Among different experimental animals, rodents are the most commonly used models in neuroscience~\cite{ellenbroek2016rodent}, primarily due to their genetic and neuroanatomical resemblance to humans~\cite{bryda2013mighty}. This resemblance makes rodents suitable subjects for the investigation of neural activity and behavioral functioning. In particular, rodent behavior provides important insights into neural processes and social functioning~\cite{ko2017neuroanatomical}. Behavioral patterns such as approaching, following, and social contact are frequently analyzed in order to better understand social communication and emotional responses~\cite{ko2017neuroanatomical}. Moreover, the observation of behavioral alterations enables researchers to infer the effects of stress, disease conditions, and pharmacological interventions~\cite{popik2024effects}. Therefore, the analysis of rodent social behavior plays a significant role in neuroscience as well as pharmacological research~\cite{haanell2014structured}.

Traditional methods for the analysis of rodent social behavior mainly rely on the manual observation and scoring of recorded videos~\cite{desland2014manual}. However, this process is labor-intensive and is more prone to human error, particularly in those situations where social interactions occur rapidly and involve overlapping body movements that are difficult to distinguish visually~\cite{gulinello2019rigor}. While subtle body movements may  contain important behavioral cues,  such cues can easily be overlooked during the manual annotation process~\cite{geros2020improved}. Therefore, these limitations negatively affect the consistency and reproducibility across different studies.

To address these challenges, early automated approaches relied on handcrafted features derived from body position, movement, and orientation~\cite{von2021big}. Although these methods helped advance automated behavior analysis, they struggled to capture complex social interactions such as nose-to-nose contact, following, and aggressive encounters under varying experimental conditions~\cite{saoud2024beyond}.

More recently, deep learning-based methods have been increasingly applied to recognize behaviors directly from video data. In this regard, convolutional neural networks (CNNs) and recurrent neural networks (RNNs) have been widely utilized to capture the spatial and temporal patterns present in behavioral sequences. For instance, DeepEthogram~\cite{bohnslav2021deepethogram} introduced a CNN-based pipeline for frame-level behavior detection, whereas MARS~\cite{segalin2021mouse} combined pose estimation with classification in order to analyze mouse social interactions more effectively.

In behavior recognition, two main input representations are commonly used. Pixel-based approaches process raw video frames and capture visual appearance, but they are sensitive to different settings, lighting, background, and camera conditions. Pose-based approaches instead rely on keypoint coordinates representing the body positions of animals. These methods focus on motion patterns and are generally more robust to visual variations, although they depend on the accuracy of the pose estimation stage.

\begin{figure}[t]
\centering
\includegraphics[width=0.85\textwidth]{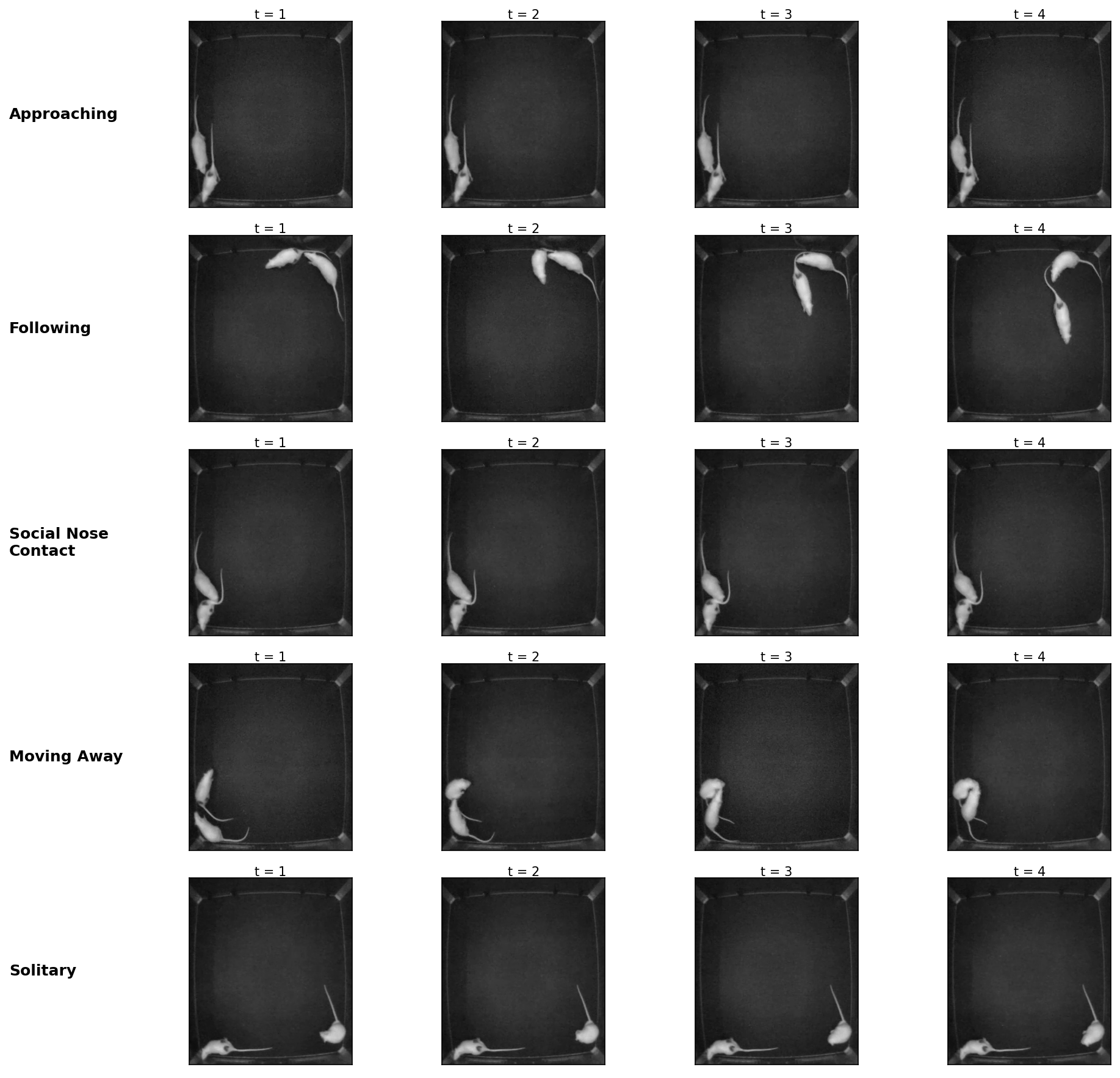}
\caption{Four consecutive frames for each of the five behaviors in the RatSI~\cite{lorbach2018learning} dataset, showing how each behavior develops over time. Behavior instance durations vary across videos, with \textit{Solitary} and \textit{Following} typically spanning longer periods and \textit{Moving Away} occurring over shorter spans.}
\label{fig:behavior_sequences}
\end{figure}

In spite of the recent advancements, the modeling of rodent 
behavior is still considered a challenging task because distinct 
behaviors are exhibited at different temporal scales~\cite{kuo2022using, 
van2023disentangling}. As illustrated in Figure~\ref{fig:behavior_sequences}, behaviors such as \textit{Approaching} and \textit{Following} are performed over multiple frames with continuous full body motion, whereas \textit{Moving Away} tends to occur over shorter temporal spans, reflecting brief disengagement episodes within the  sequence. Therefore, the accurate recognition of both slow and fast behaviors requires models that are capable of learning motion dynamics at multiple temporal scales~\cite{fazzari2025animal}.

A major challenge, therefore, lies in modeling behaviors that occur over substantially different temporal durations within the same interaction sequence. Certain events may persist for only a few frames and are characterized by subtle pose variations, whereas other behaviors emerge progressively over longer temporal intervals. Models that primarily focus on local temporal information may fail to capture long-range interaction dependencies, while approaches that emphasize global sequence-level context may overlook short but behaviorally significant motion patterns. This multi-scale temporal characteristic of rodent behavior thus motivates the development of architectures capable of simultaneously capturing fine-grained local motion cues and longer-term behavioral dynamics.

Transformers have recently shown strong potential for video understanding due to their ability to model long-range dependencies using attention mechanisms~\cite{fish2022two, vaswani2017attention}. Pose-based behavioral data naturally forms a temporal sequence of structured coordinate vectors, making it well suited for attention-based sequence modeling. Unlike recurrent architectures that process frames sequentially and may struggle with long-range dependencies, transformer attention mechanisms allow each frame to directly attend to all other frames in the sequence. This property enables the model to capture complex temporal relationships between motion patterns that may occur at distant points in time within an interaction sequence.

Most transformer architectures emphasize global context and may overlook short and localized motion cues~\cite{sharif2025rodent, liu2021swin, arnab2021vivit}. This limitation is particularly problematic in rodent behavior recognition. The reason is that many behaviors appear similar at a coarse level and differ only in brief and subtle movements. For example both \textit{Approaching} and \textit{Following} involve forward body motion while \textit{Social Nose Contact} is distinguished by a short head movement that spans only a few frames. Due to this reason a model that focuses solely on global sequence patterns may confuse these behaviors.

Dataset imbalance is another important challenge in the process of rodent behavior analysis. The reason is that rodent behavior datasets are mostly dominated by long duration behaviors such as solitary activity whereas short but important behaviors such as pinning and nape attacking occur relatively rarely. Due to this reason the models learn the dominant classes in a more effective way. However the detection of rare but behaviorally significant events becomes a difficult task for these models~\cite{lorbach2018learning}.

To address these challenges, we propose MSGL-Transformer, a multi-scale global-local Transformer designed for rodent social behavior recognition. The model integrates fine-grained local motion cues with long-range temporal context to capture behavioral dynamics across multiple time scales. Together, these design choices enable the model to effectively capture both subtle short-term movements and longer interaction patterns in rodent social behavior sequences. 

The proposed MSGL-Transformer is based on three important design elements. The first important element is the multi-scale temporal attention mechanism. The method works by jointly analyzing short-range, medium-range and global temporal dependencies. Another important element is the Behavior-Aware Modulation (BAM) block. This block works by dynamically reweighting the temporal embeddings in order to emphasize the motion patterns which are more informative for behavior analysis. Third, the model makes direct use of the pose coordinate sequences, unlike prior skeleton based transformer approaches which are dependent on predefined spatial graphs or hierarchical temporal windowing. Due to this reason the architecture is lightweight and flexible. Another advantage achieved through this method is that it can generalize across different datasets that may exhibit different pose configurations.

We evaluate the proposed model on two publicly available datasets: RatSI~\cite{lorbach2018learning}, which contains a diverse set of rat social behaviors, and CalMS21~\cite{sun2021multi}, a large-scale mouse dyadic interaction dataset with a different species, keypoint layout, and behavior vocabulary. To summarize, the main contributions of this work are:

\begin{itemize}
    \item We propose MSGL-Transformer, a transformer-based model for recognizing rodent social behaviors. The model captures both short-term and long-term temporal patterns through multi-scale global-local attention, enabling effective recognition of rapid actions and longer behavioral interactions.
    
    \item We introduce a Behavior-Aware Modulation (BAM) block inspired by the channel reweighting mechanism in Squeeze-and-Excitation (SE) Networks~\cite{hu2018squeeze}. Unlike SE, which operates on channel features, BAM modulates temporal embeddings to emphasize behavior-relevant dynamics.
    
    \item Experiments on the RatSI dataset~\cite{lorbach2018learning} show that MSGL-Transformer outperforms baseline models including TCN, LSTM, and Bi-LSTM across nine cross-validation splits.
    
    \item The proposed architecture generalizes across datasets without structural modifications. On CalMS21~\cite{sun2021multi}, the model achieves 87.1\% accuracy and surpasses the best-performing published keypoint-based method, HSTWFormer~\cite{ru2024hierarchical}, by +10.7\% in average per-class accuracy. It also outperforms other skeleton-based methods including 
ST-GCN~\cite{yan2018spatial}, MS-G3D~\cite{liu2020disentangling}, 
CTR-GCN~\cite{chen2021channel}, and STGAT~\cite{huang2020stgat}, 
as reported in~\cite{ru2024hierarchical}.
\end{itemize}

\section{Related Work}

Animal behavior recognition is a challenging task in the domain of automated behavior analysis. Several methods are presented in the literature for this task, where the initial techniques are based on handcrafted features and the recent methods are based on deep learning models that learn spatial and temporal patterns directly from the data. In this section, a review of prior work is presented by focusing on handcrafted approaches, deep learning methods, pose-based representations, and recent transformer-based models.

\subsection{Handcrafted Behavior Recognition}

Early automated methods for animal behavior analysis primarily used 
handcrafted visual and motion features derived from video 
recordings~\cite{anderson2014toward, dechaumont2012computerized}. Jhuang et al.~\cite{yu2009automated} developed an early framework that combined background subtraction with shape based descriptors for rodent behavior analysis. This system was able to recognize behaviors such as walking, grooming, and rearing under controlled single animal conditions. However, it remained sensitive to lighting variation and background noise. Overall, these results illustrate the limited robustness of handcrafted video based approaches under changing recording conditions.

Giancardo et al.~\cite{giancardo2013automatic} introduced a trajectory-based method that extracted motion statistics from tracking data to classify social behaviors between pairs of mice, including following and approaching. Burgos Artizzu et al.~\cite{burgos2012social} further incorporated engineered spatial features such as relative distance, orientation, and velocity between animals to recognize social behaviors such as attack, investigation, and mounting. 

Although these approaches demonstrate the feasibility of automated behavior recognition, they mainly rely on manually designed features and remain limited in their capability to effectively model complex temporal dynamics.

\subsection{Deep Learning Approaches for Behavior Recognition}

Deep learning methods are some of the most effective and recent 
approaches used for the automatic analysis of behavior from raw 
video data~\cite{fazzari2025animal, mathis2020deep}. The reason for their effectiveness is that these methods are capable of learning representative features directly from the input data without the need of hand crafted extraction procedures. In this regard, convolutional neural networks~\cite{lecun1998gradient} and recurrent neural networks~\cite{hochreiter1997long} are commonly used. Convolutional neural networks mainly help in capturing the spatial patterns, whereas recurrent neural networks are useful for analyzing temporal patterns in behavioral sequences. 

For example, DeepEthogram~\cite{bohnslav2021deepethogram} introduced a CNN-based pipeline for recognizing animal behaviors directly from video frames without requiring pose estimation. Strong performance is achieved through the method for behaviors such as grooming and locomotion. Similarly, Kuo et al.~\cite{kuo2022using} applied an RNN-based approach to model temporal dynamics in rodent behavior sequences, demonstrating that recurrent architectures can capture short-term behavioral transitions. However, video based approaches are sensitive to background clutter, lighting variations and camera viewpoint changes. These factors may reduce the generalization capability of the method across different recording environments.

\subsection{Pose-Based Behavior Recognition}

Pose-based approaches operate on keypoint coordinates which represent the spatial positions of different body parts and are used for behavior analysis when the focus is on the movement pattern of body parts instead of the raw 
appearance information~\cite{chen2025twostream, unsupervised2025pose}. By employing  keypoints, the representation becomes less sensitive to background clutter, lighting variations and camera viewpoint changes. 

The MARS approach proposed by Segalin et al.~\cite{segalin2021mouse} falls into this category. The framework  combines pose estimation with behavior classification in order to recognize mouse social behaviors such as attack, mounting and close investigation.  Nilsson et al.~\cite{nilsson2020simple} introduced SimBA,  a supervised learning toolkit that enables behavior classification through pose derived features. 
An unsupervised approach is also proposed by Luxem et al.~\cite{luxem2022identifying} where deep variational embeddings are used to discover latent temporal patterns in animal motion. 

Although pose-based approaches improve the robustness against visual variations, many of the existing methods still depend on handcrafted features or limited temporal modeling.

\subsection{Transformer-Based Behavior Recognition}

Transformers~\cite{vaswani2017attention} have recently emerged as a powerful architecture for modeling long-range dependencies in sequential data. Vision-based transformers such as ViT~\cite{dosovitskiy2020image} and Swin Transformer~\cite{liu2021swin} demonstrated strong performance in image and video understanding by leveraging self-attention mechanisms. More recently, Wang et al.~\cite{wang2026animal} 
extended this direction to animal behavioral analysis using self-supervised pretraining of vision transformers on unlabeled video.

Transformer architectures have been applied to pose-based animal behavior recognition. Ru and Duan~\cite{ru2024hierarchical} proposed HSTWFormer, a hierarchical spatial–temporal window transformer that captures correlations between keypoints across frames. The model achieved strong performance on the CRIM13 and CalMS21 datasets and outperformed several skeleton-based baselines including ST-GCN, MS-G3D, CTR-GCN, and STGAT.

Despite these advances, existing models still face challenges in capturing both fine-grained short-term motion cues and long-range temporal dependencies in rodent social behaviors.

\section{Materials and Methods}

\subsection{Data Description}
\label{sec:datasets}

To evaluate the generalizability of the proposed approach, we use two publicly available datasets: RatSI~\cite{lorbach2018learning} and CalMS21~\cite{sun2021multi}. The datasets differ in terms of species they cover (rats vs.\ mice), number of keypoints, input dimensionality (12 vs.\ 28), and behavior vocabularies (5 vs.\ 4 classes). The two datasets are described below, followed by a brief comparison of the two datasets.

\subsubsection{RatSI Dataset}
The Rat Social Interaction (RatSI) dataset~\cite{lorbach2018learning} contains video recordings of dyadic interactions between two rats. The dataset consists of nine observation videos of approximately 15 minutes each (Observation01 to Observation09), where each video represents an independent interaction session. For each frame, pose data is provided in the form of two-dimensional $(x,y)$ coordinates for six keypoints: the nose, center, and tailbase of both the male and female rats, resulting in a 12-dimensional pose representation ($3$ keypoints $\times$ $2$ rats $\times$ $2$ coordinates).

Each frame is annotated with one of several behavior labels by domain experts, including \textit{Solitary}, \textit{Approaching}, \textit{Following}, \textit{Moving Away}, \textit{Social Nose Contact}, \textit{Allogrooming}, \textit{Nape Attacking}, \textit{Pinning}, \textit{Other}, and \textit{Uncertain}. The dataset contains approximately 202,550 labeled frames across all observation videos. Table \ref{tab:ratsi_distribution} shows the distribution of behavior classes across the nine observation videos. 

\begin{table*}[t]
\centering
\caption{Frame distribution of social behaviors across all nine RatSI observation videos. The last row indicates the overall percentage of each behavior across all videos (total 202{,}550 frames).}
\resizebox{\textwidth}{!}{
\begin{tabular}{lrrrrrrrrrrr}
\toprule
\textbf{Video} & 
\textbf{Solitary} & 
\textbf{Approaching} & 
\textbf{Following} & 
\textbf{Moving Away} & 
\textbf{\makecell{Social\\Nose\\Contact}} &\textbf{Allogrooming} & 
\textbf{Nape Attacking} & 
\textbf{Pinning} & 
\textbf{Other} & 
\textbf{Uncertain} & 
\textbf{Total Frames} \\
\midrule
Observation01 & 12{,}281 & 1{,}405 & 2{,}150 & 985 & 1{,}799 & 1{,}778 & 221 & 16 & 792 & 88 & \textbf{21{,}515} \\
Observation02 & 14{,}709 & 1{,}066 & 593 & 592 & 1{,}754 & 2{,}240 & 95 & 0 & 231 & 55 & \textbf{21{,}335} \\
Observation03 & 15{,}428 & 1{,}083 & 2{,}522 & 482 & 1{,}106 & 587 & 104 & 1{,}036 & 472 & 21 & \textbf{22{,}841} \\
Observation04 & 14{,}776 & 1{,}528 & 444 & 931 & 2{,}621 & 2{,}003 & 141 & 0 & 300 & 0 & \textbf{22{,}744} \\
Observation05 & 14{,}297 & 1{,}310 & 2{,}955 & 880 & 1{,}067 & 111 & 434 & 127 & 1{,}678 & 0 & \textbf{22{,}859} \\
Observation06 & 13{,}152 & 2{,}443 & 2{,}141 & 1{,}503 & 1{,}980 & 486 & 192 & 0 & 856 & 0 & \textbf{22{,}753} \\
Observation07 & 11{,}405 & 2{,}058 & 3{,}359 & 1{,}202 & 1{,}865 & 826 & 581 & 30 & 1{,}598 & 0 & \textbf{22{,}924} \\
Observation08 & 12{,}323 & 2{,}066 & 2{,}696 & 1{,}391 & 3{,}422 & 579 & 108 & 0 & 188 & 0 & \textbf{22{,}773} \\
Observation09 & 11{,}295 & 2{,}259 & 2{,}017 & 936 & 4{,}141 & 799 & 103 & 15 & 1{,}241 & 0 & \textbf{22{,}806} \\
\midrule
\textbf{All Videos (\%)} & 
\textbf{59.11} & 
\textbf{7.52} & 
\textbf{9.32} & 
\textbf{4.40} & 
\textbf{9.76} & 
\textbf{4.65} & 
\textbf{0.98} & 
\textbf{0.60} & 
\textbf{3.63} & 
\textbf{0.08} & 
\textbf{202{,}550} \\
\bottomrule
\end{tabular}
}
\label{tab:ratsi_distribution}
\end{table*}

\begin{figure}[H]
    \centering
    \includegraphics[width=\linewidth]{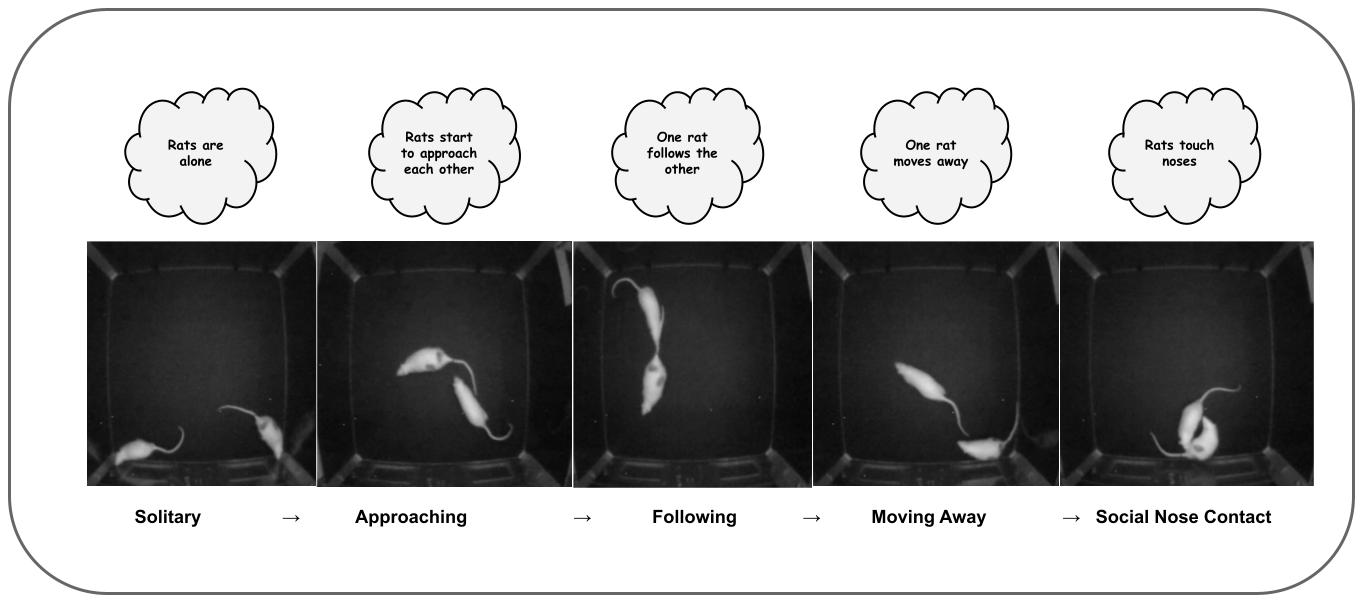}
    \caption{Sample frames from the RatSI dataset showing the five 
social behaviors studied in this work: Solitary, Approaching, 
Following, Moving Away, and Social Nose Contact, spanning from 
no social engagement to direct physical contact.}
    \label{fig:rats_interaction}
\end{figure}

In this work, we focus on five behaviors: \textit{Solitary}, \textit{Approaching}, \textit{Following}, \textit{Moving Away}, and \textit{Social Nose Contact}. These behaviors account for approximately 90\% of all labeled frames and represent the primary range of social interaction, from isolation to direct physical contact. Table~\ref{tab:behaviors} summarizes these behaviors, while Figure~\ref{fig:rats_interaction} illustrates example frames.

\begin{table}[ht]
 \small
\centering
\caption{Summary of the selected RatSI behaviors that are considered to represent the core range of rodent social interactions.}
\begin{tabular}{ll}
\toprule
\textbf{Behavior} & \textbf{Description / Social Context} \\
\midrule
\textbf{Solitary} & The rat is \textit{alone}, showing no social engagement (isolation phase). \\
\textbf{Approaching} & The rat \textit{initiates} interaction by moving toward another (beginning of engagement). \\
\textbf{Following} & Indicates \textit{sustained interaction}, maintaining social attention. \\
\textbf{Moving Away} & Shows \textit{withdrawal or avoidance}, marking disengagement. \\
\textbf{Social Nose Contact} & Reflects \textit{direct physical interaction}, the highest level of social engagement. \\
\bottomrule
\end{tabular}
\label{tab:behaviors}
\end{table}

Rare behaviors such as \textit{Pinning} and \textit{Nape Attacking} were excluded because they occur infrequently and introduce severe class imbalance. The distribution of all annotated behaviors across the nine observation videos is shown in Table~\ref{tab:ratsi_distribution}. Most frames correspond to the \textit{Solitary} class (59.1\%), followed by \textit{Social Nose Contact} (9.8\%) and \textit{Following} (9.3\%).

\subsubsection{CalMS21 Dataset}
To evaluate cross-dataset generalization, we additionally test the proposed MSGL-Transformer on the Caltech Mouse Social Interactions dataset (CalMS21)~\cite{sun2021multi}. CalMS21 contains top-view recordings of pairs of freely interacting mice, with pose keypoints extracted using the MARS tracking framework~\cite{segalin2021mouse}.
Each mouse is represented by seven keypoints (nose, left ear, right ear, neck, left hip, right hip, and tail base), resulting in a 28-dimensional pose representation ($7$ keypoints $\times$ $2$ mice $\times$ $2$ coordinates). We use Task~1 of the dataset, which contains 769,845 labeled frames across 89 videos (including 70 training and 19 test videos).

\begin{table}[ht]
\small
\centering
\caption{Summary of CalMS21 behaviors representing the 
core range of mouse dyadic social interactions.}
\begin{tabular}{ll}
\toprule
\textbf{Behavior} & \textbf{Description / Social Context} \\
\midrule
\textbf{Attack} & \textit{Aggressive contact} initiated by 
the resident mouse toward the intruder. \\
\textbf{Investigation} & The resident mouse \textit{closely 
sniffs or investigates} the intruder mouse. \\
\textbf{Mount} & The resident mouse \textit{mounts} the 
intruder, reflecting dominance behavior. \\
\textbf{Other} & All \textit{remaining frames} not 
belonging to the three active social behaviors. \\
\bottomrule
\end{tabular}
\label{tab:calms21_behaviors}
\end{table}

\begin{figure}[h]
    \centering
    \includegraphics[width=\textwidth]{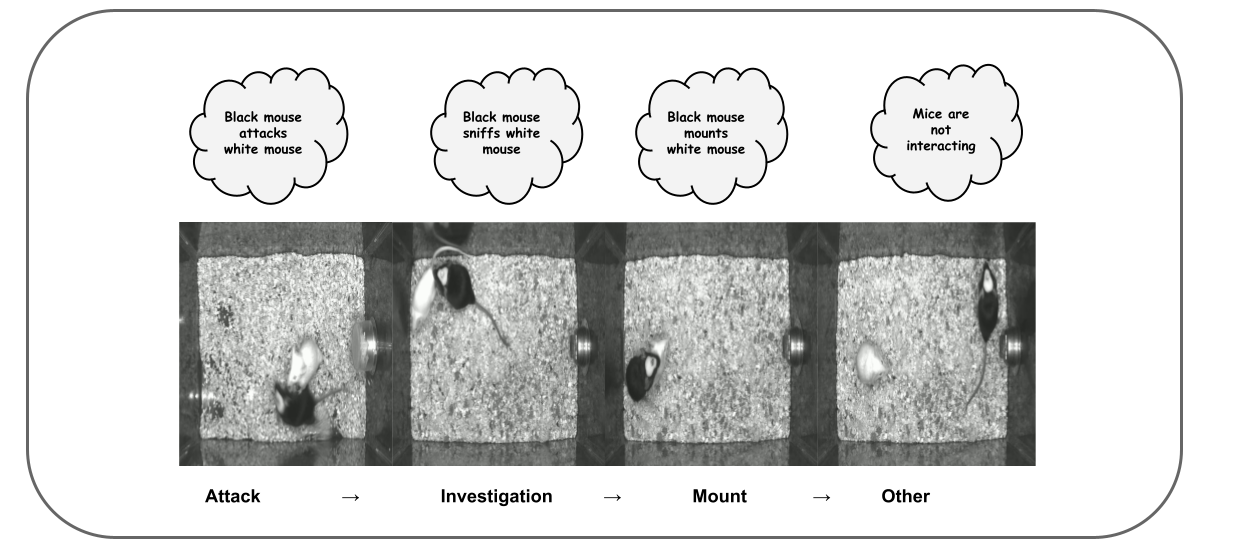}
    \caption{Representative video frames from the CalMS21 dataset illustrating 
    the four annotated behavior classes. The resident (black) mouse initiates 
    all active behaviors toward the intruder (white) mouse in the 
    resident-intruder assay~\cite{sun2021multi}.}
    \label{fig:calms21_behaviors}
\end{figure}

\begin{table}[ht]
\small
\centering
\caption{Frame distribution of behavior classes across 
training and test splits of the CalMS21 Task 1 dataset 
(total 769,845 frames across 89 videos).}
\begin{tabular}{lrrrrr}
\toprule
\textbf{Split} & \textbf{Attack} & \textbf{Investigation} 
& \textbf{Mount} & \textbf{Other} & \textbf{Total Frames} \\
\midrule
Train (70 videos) & 14,039 & 146,615 & 28,615 
& 318,469 & \textbf{507,738} \\
Test (19 videos)  & 12,630 &  61,275 & 31,848 
& 156,354 & \textbf{262,107} \\
\midrule
\textbf{All Videos} & \textbf{26,669} & \textbf{207,890} 
& \textbf{60,463} & \textbf{474,823} & \textbf{769,845} \\
\textbf{Train (\%)} & \textbf{2.76} & \textbf{28.88} 
& \textbf{5.64} & \textbf{62.73} & — \\
\bottomrule
\end{tabular}
\label{tab:calms21_distribution}
\end{table}

The four behavior classes in the CalMS21 dataset are \textit{Attack}, \textit{Investigation}, \textit{Mount}, and \textit{Other}, as summarized in Table~\ref{tab:calms21_behaviors}. Similar to RatSI, the dataset exhibits class imbalance, with \textit{Other} being the most frequent class (62.7\%) and \textit{Attack} the rarest (2.8\% of the training data). Figure~\ref{fig:calms21_behaviors} shows representative frames 
for each of the four behavior classes in the CalMS21 dataset.
Table \ref{tab:calms21_distribution} shows the distribution of the behavior classes across the train and test subsets. The largest {\it Other} class represents close to 62.73\% of the train subset, while the smallest classes represents only 2.76\% ({\it Attack}) and 5.64\% ({\it Mount}).  Similar to the RatSI dataset, these statistics show that the  CalMS21 dataset exhibits severe class imbalance.

While there are some similarities between the RatSI and CalMS21 datasets, Table~\ref{tab:dataset_comparison} summarizes their key differences. Despite these differences, the same model architecture and hyperparameters are used for both datasets, with only the input dimensionality ($D$) and number of output classes ($C$) adapted.

\begin{table}[ht]
\small
\centering
\caption{Comparison of the two evaluation datasets.}
\begin{tabular}{lcc}
\toprule
\textbf{Aspect} & \textbf{RatSI} & \textbf{CalMS21} \\
\midrule
Species & Rats & Mice \\
Number of Videos & 9 & 89 \\
Total Labeled Frames & \textasciitilde202,550 & \textasciitilde769,845 \\
Keypoints per Animal & 3 & 7 \\
Input Dimension ($D$) & 12 & 28 \\
Behavior Classes ($C$) & 5 & 4 \\
Dominant Class & Solitary (59.1\%) & Other (62.7\%) \\

Rarest Class & Moving Away (4.4\%) & Attack (2.8\%) \\

\bottomrule
\end{tabular}
\label{tab:dataset_comparison}
\end{table}

\subsection{Problem Formulation}
\label{sec:problem_setup}

Given a dataset consisting of pose trajectories extracted from dyadic rodent interactions, where each frame is associated with a behavior label, rodent social behavior recognition is formulated as a temporal sequence classification problem. More specifically, given a sequence of pose vectors extracted from interacting animals, the goal is to predict the behavior label associated with the sequence.

At each time step $t$, the pose is represented by a feature vector
$x_t \in \mathbb{R}^{D},$
where $D$ denotes the dimensionality of the pose representation. Each vector contains the two-dimensional coordinates of key body points for the two interacting animals.

From the pose sequence, we construct overlapping temporal windows of length $T$. Each training sample $X_i$ is therefore defined as
$
X_i = [x_i, x_{i+1}, \ldots, x_{i+T-1}] \in \mathbb{R}^{T \times D}.
$

The model uses the full temporal window as context and predicts the behavior label corresponding to the final frame in the sequence.  Specifically, the label $y_i$ of the instance $X_i$
is the label of the frame at position $i+T-1$ in the sequence corresponding to $X_i$, i.e., $y_i = y_{i+T-1}.$ The final-frame labeling strategy is commonly used in behavior recognition tasks because it preserves temporal causality and allows the model to use preceding frames as context for predicting the current behavior.

The temporal window length is $T=35$ frames. During training, a sliding window with stride $1$ is used to generate overlapping sequences/instances. Thus, consecutive windows/sequences overlap by 34 frames. Even if one window starts at the end of a behavior, the following windows will gradually shift into the next behavior. This formulation allows the model to capture temporal dependencies between frames and learn how behaviors evolve over time.

Formally, the behavior recognition model is defined as a function
$
f_{\theta} : \mathbb{R}^{T \times D} \rightarrow \mathbb{R}^{C},
$
where $C$ denotes the number of behavior classes. The predicted class probabilities are computed using the softmax function
$
\displaystyle\hat{p}_c = \frac{e^{f_{\theta}(X)_c}}{\sum_k e^{f_{\theta}(X)_k}},
$
and the final predicted label is
$
\hat{y} = \arg\max_c \hat{p}_c .
$

\subsection{Data Preprocessing and Pipeline}

Pose coordinate sequences extracted from the datasets described in Section~\ref{sec:problem_setup} are processed through a fixed preprocessing pipeline before being used for training the model. The pipeline consists of four main steps: missing value imputation, feature normalization, temporal window generation, and tensor conversion.

\subsubsection{Filling-in Missing Values}  
Some pose coordinates contain missing values due to tracking failures. Missing values are handled using mean imputation applied independently to each coordinate feature~\cite{pedregosa2011scikit}. For a feature $j$, the imputed value is computed as
\[
x'_{t,j} =
\begin{cases}
x_{t,j}, & \text{if the value is valid} \\
\frac{1}{N_j}\sum_{n=1}^{N_j} x_{n,j}, & \text{otherwise}
\end{cases}
\]
where $N_j$ is the number of valid observations for feature $j$ across the training set. The mean value obtained from the training set is also used to handle missing values in the test and validation subsets.

\subsubsection{Feature Normalization}  
After imputation, all features are standardized to zero mean and unit variance:
$\displaystyle
x^{\text{norm}}_{t,j} = \frac{x'_{t,j} - \mu_j}{\sigma_j},
$
where $\mu_j$ and $\sigma_j$ denote the mean and standard deviation of feature $j$ computed from the training subset. The same values are used to normalize the features in the validation/test subsets.

\subsubsection{Temporal Window Construction}  
From the normalized pose sequences, overlapping temporal samples are generated using a sliding window of length $T = 35$ and stride $1$. Each window
$
X_i = [x_i, x_{i+1}, \ldots, x_{i+T-1}]
$
forms a training sample, with the corresponding label $y_i$ defined as the behavior label of the final frame in the window $(i + T - 1)$. This formulation preserves temporal continuity and substantially increases the number of training samples. There is no overlap between sequences in the train and validation/test subsets. 

\subsubsection{Tensor Conversion}  
All generated sequences are converted into PyTorch tensors~\cite{paszke2019pytorch} and grouped into mini-batches of 32 samples for efficient GPU training.

The identical preprocessing pipeline is applied to both RatSI and CalMS21 datasets. The only difference is the dimensionality of the pose representation: $D=12$ for RatSI and $D=28$ for CalMS21. This consistent pipeline converts raw coordinate trajectories into standardized temporal tensors that serve as input to the proposed MSGL-Transformer.

To ensure reproducibility, all preprocessing components (imputer, scaler, and label encoder) are stored using \texttt{joblib}. 

\subsection{Model Architecture}

\begin{figure}[t]
    \centering
    \includegraphics[width=0.9\linewidth]{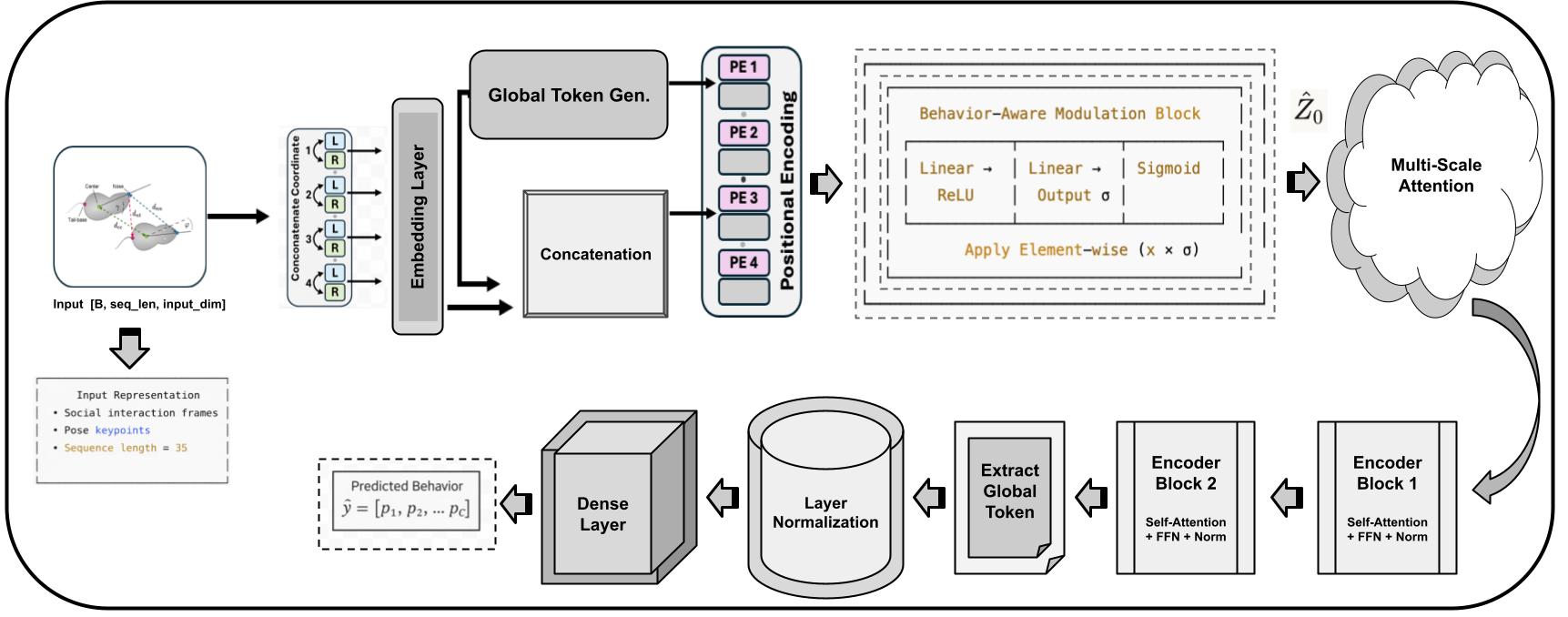}
    \caption{Overview of the proposed MSGL-Transformer architecture. The model combines global token generation, behavior-aware modulation, and multi-scale attention within a lightweight transformer encoder to jointly capture global context and fine-grained temporal dynamics in rodent social interactions.}
    \label{fig:proposed_method}
\end{figure}

The proposed Multi-Scale Global-Local Transformer (MSGL-Transformer) processes a temporal pose sequence 
$\mathbf{X} \in \mathbb{R}^{T \times D}$ and predicts one of $C$ behavior classes.   An overview of the architecture is shown in Figure~\ref{fig:proposed_method}.
The architecture consists of five main components: (1) an input embedding with a learnable global token, 
(2) a behavior-aware modulation BAM block inspired by Squeeze-and-Excitation networks~\cite{hu2018squeeze}, 
(3) a multi-scale attention module, (4) a lightweight transformer encoder~\cite{vaswani2017attention}, 
and (5) a classification head. The components of the model are described in what follows.

\subsubsection{Input Embedding and Global Token}

Each input sequence $\mathbf{X} \in \mathbb{R}^{T \times D}$ consists of $T=35$ pose vectors, where each vector 
$\mathbf{x}_t \in \mathbb{R}^{D}$ represents the coordinates of key body points for the two interacting animals. 
The pose coordinates are linearly projected to a model dimension $d=64$ using an embedding matrix 
$\mathbf{W}_e \in \mathbb{R}^{D \times d}$.

A learnable global token $\mathbf{g}_0 \in \mathbb{R}^{1 \times d}$ is prepended to the embedded input sequence and serves as a summary representation of the entire sequence, similar to the \texttt{[CLS]} token used in Vision Transformers~\cite{dosovitskiy2020image}. 
The concatenated sequence of the global token and the embedded pose vectors is combined with a learnable positional encoding $\mathbf{P} \in \mathbb{R}^{(T+1) \times d}$, yielding the augmented representation

\subsubsection{Behavior-Aware Modulation Block}

The BAM block introduces a global modulation mechanism that adapts feature scaling according to the behavioral context. Inspired by the channel reweighting strategy used in SE networks~\cite{hu2018squeeze}, BAM learns a modulation vector from the entire pose sequence and applies it to the embedded features.

The non-global tokens of $\mathbf{Z}_0$ are flattened and passed through a two-layer fully connected network with ReLU and Sigmoid activations:
$
m = \sigma\!\left(
\text{Linear}\left(
\text{ReLU}\left(
\text{Linear}\left(\text{vec}(\mathbf{Z}_0[:,1:,:])\right)
\right)
\right)
\right),
$
where $m \in \mathbb{R}^{d}$ is a modulation vector. The features are then rescaled channel-wise:
$
\hat{\mathbf{Z}}_0 = \mathbf{Z}_0 \odot m .
$
This operation emphasizes behavior-relevant motion patterns while suppressing less informative features.

\subsubsection{Multi-Scale Attention Module}

\begin{figure}[t]
\centering
\includegraphics[width=0.75\textwidth]{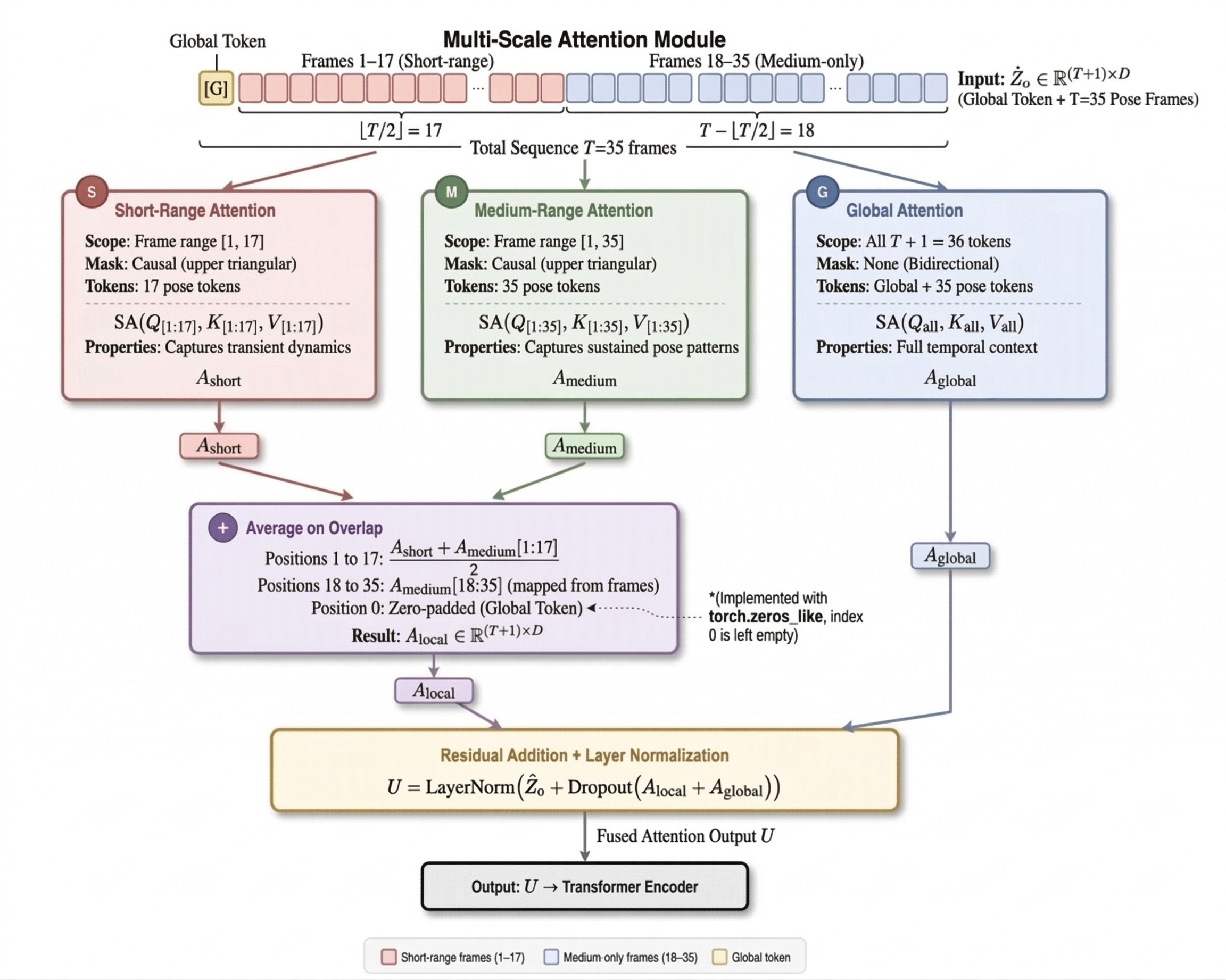}
\caption{Structure of the multi-scale attention module. The modulated 
input $\hat{Z}_0$ is passed through three parallel branches: short-range 
causal attention on the first $\lfloor T/2 \rfloor$ frames (this branch focuses on rapid motion cues occurring within short temporal contexts), medium-range causal attention 
on all $T$ frames, and global bidirectional attention on all $T+1$ tokens. The short-range and medium-range outputs are averaged on overlapping positions to produce $A_{local}$, which is combined with $A_{global}$ through residual addition and layer normalization.}
\label{fig:msa}
\end{figure}

The multi-scale attention module captures temporal dependencies 
across different time ranges, inspired by multi-scale modeling 
strategies in video understanding~\cite{fan2021multiscale, chen2021multi}, 
as shown in Figure~\ref{fig:msa}. The module consists of two local 
attention branches and one global attention branch.

The short-range branch is employed to process the first $\lfloor T/2 \rfloor$ frames using a causal masking strategy, where each frame is allowed to attend only to itself and the previous frames. Thereafter, the medium-range branch processes all $T$ frames by utilizing the same masking strategy. Therefore, these two branches are helpful to model the short-term and medium-term temporal dynamics more effectively.

Since the first $\lfloor T/2 \rfloor$ frames are processed by both branches, the corresponding outputs within this region are averaged in order to generate the local attention representation $A_{\text{local}}$. 

The global branch operates on all $T+1$ tokens, including the global token, using full bidirectional attention without masking. This branch captures long-range dependencies and produces the representation $A_{\text{global}}$.
The outputs are combined through residual addition followed by layer normalization:
$
U = \text{LayerNorm}(\hat{\mathbf{Z}}_0 + 
\text{Dropout}(A_{\text{local}} + A_{\text{global}})).
$
This design enables the model to jointly capture fine-grained motion patterns and long-range behavioral context. 

\subsubsection{Transformer Encoder and Classification Head}

The resulting representation is further processed by a lightweight transformer encoder consisting of two layers with feed-forward dimension $d_{ff}=128$. Each encoder layer follows the standard transformer structure:
$U' = \text{LayerNorm}(U + \text{MHA}(U)),$
$
V = \text{LayerNorm}(U' + \text{FFN}(U')) .
$
After the final encoder layer, the global token $v_g = V[0]$ is extracted and passed through a dropout layer ($p=0.2$) followed by a linear classifier to produce logits
$
f_\theta(X) \in \mathbb{R}^{C}.
$
Class probabilities are obtained using the softmax function. 

\subsection{Implementation Details} 
\begin{table}[t]
\small
\centering
\caption{Summary of model and training hyperparameters used in the proposed MSGL-Transformer.}
\begin{tabular}{l c c l}
\hline
\textbf{Hyperparameter} & \textbf{Symbol / Setting} & \textbf{Value} & \textbf{Description} \\
\hline
Sequence length & $T$ & 35 & Number of time steps per sequence \\
Input dimension & $D$ & 12 (RatSI) / 28 (CalMS21) & Coordinate features per frame \\
Output classes & $C$ & 5 (RatSI) / 4 (CalMS21) & Number of behavior classes \\
Model dimension & $d_{\text{model}}$ & 64 & Embedding size for each token \\
Feed-forward dimension & $d_{\text{ff}}$ & 128 & Hidden size of the FFN  \\
Attention heads & $h$ & 4 & Number of heads \\
Encoder layers & $L$ & 2 & Depth of transformer encoder \\
Dropout rate & $p$ & 0.2 & Applied uniformly to all major layers \\
Batch size & – & 32 & Training mini-batch size \\
Optimizer & – & Adam & Learning rate $=0.001$ \\
Learning rate scheduler & – & ReduceLROnPlateau & Factor $=0.5$, patience $=5$ epochs \\
Early stopping & – & Enabled & Patience $=25$ epochs \\
Loss function & – & Label-smoothing CE & Smoothing $=0.1$ \\
Max epochs & – & 50 & Maximum number of training epochs \\
\hline
\end{tabular}
\label{tab:hyperparams}
\end{table}

All model weights are initialized using Xavier uniform initialization with zero bias. Model hyperparameters are summarized in Table~\ref{tab:hyperparams}. 
To validate the choice of the temporal window length $T=35$,  we fine-tuned the window size on the CalMS21  dataset. The results for three window lengths ($T=20$, $T=35$ and $T=50$) are shown   in Table~\ref{tab:window_ablation}. All evaluated window sizes achieved similar  accuracy (approximately 87\%). However, the configuration with $T = 35$ provided improved per-class performance for sustained behaviors such as \textit{Investigation} (F1-score: 0.788 compared with 0.773 for $T = 20$). Increasing the window size to $T = 50$ resulted in an additional 62,400 parameters without producing further performance improvement. Therefore, $T = 35$ was deemed to be a practical and efficient configuration for the proposed model.

\begin{table}[t]
\small
\centering
\caption{Effect of window size on model performance on CalMS21 Task 1.}
\label{tab:window_ablation}
\begin{tabular}{lcccc}
\hline
\textbf{Window Size} & \textbf{Parameters} & \textbf{Accuracy} & \textbf{F1-Score} & \textbf{Investigation F1} \\
\hline
$T = 20$ & 223,492 & 0.8706 & 0.8751 & 0.7728 \\
$T = 35$ & 285,892 & 0.8709 & 0.8745 & 0.7881 \\
$T = 50$ & 348,292 & 0.8711 & 0.8744 & 0.7896 \\
\hline
\end{tabular}
\end{table}

\section{Results \& Discussion}

\subsection{RatSI Results}

The proposed MSGL-Transformer was evaluated on the RatSI  dataset using a leave-one-video-out cross-validation protocol.  Unlike many benchmark datasets that provide predefined 
train/test splits, the RatSI dataset does not include an official data partition, which motivated the use of this evaluation strategy. In each split, one video was used for  testing, one for validation, and the remaining seven videos were used for training. This process was repeated across all  nine videos to assess the robustness of the model across different interaction sessions.

Table~\ref{tab:cv_results} reports the overall performance across all cross-validation splits. The model achieved a mean accuracy of $0.754 \pm 0.051$, with precision $0.750 \pm 0.044$, recall $0.754 \pm 0.051$, and F1-score $0.745 \pm 0.047$. The mean  and standard deviation  were computed over the 9 splits. 
These results indicate stable model behavior across different video splits.

\subsubsection{Quantitative and Per-Class Results}
\begin{figure}[H]
    \centering
    \includegraphics[width=0.85\linewidth]{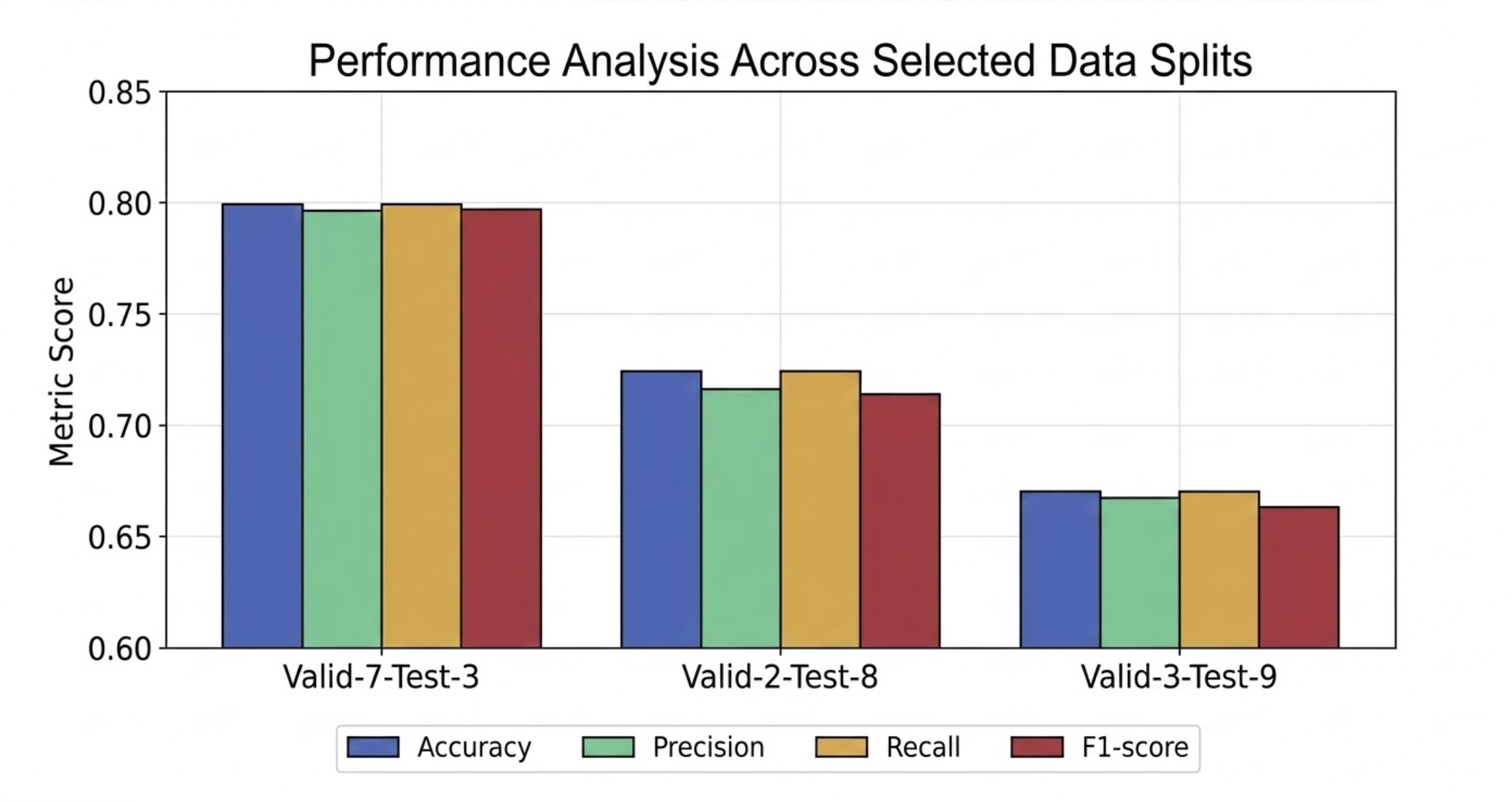}
    \caption{Accuracy, precision, recall, and F1-score across three representative splits in RatSI: Valid-7--Test-3 (best-performing), Valid-2--Test-8 (near the mean), and Valid-3--Test-9 (most challenging).}
\label{fig:three_splits_chart}
\end{figure}

To better analyze the variability in performance, three representative splits are highlighted, namely Valid-7--Test-3, Valid-2--Test-8, and Valid-3--Test-9, which correspond to the highest/best-performing case, the near-average case, and the most challenging case, respectively. Therefore, these splits represent the best, the typical, and the most difficult conditions encountered across the nine videos in the dataset.
Figure~\ref{fig:three_splits_chart} summarizes the accuracy, precision, recall, and F1-score for these three cases. 

To further analyze the model's behavior, we examined the per-class F1-scores across the same three splits (Table~\ref{tab:beh_results}). The model performs best on the \textit{Solitary} class, with F1-scores ranging from 0.798 to 0.909. This behavior dominates the dataset and has distinctive motion patterns, which likely contributes to its strong performance.
The \textit{Approaching} and \textit{Following} classes show moderate performance, with F1-scores between 0.453 and 0.595. In several cases, \textit{Following} exhibits higher recall than precision, suggesting a tendency for the model to overpredict this class.

\begin{table}[H]
\small
\centering
\caption{Cross-validation results of the MSGL-Transformer on the RatSI dataset.}
\label{tab:cv_results}
\begin{tabular}{cccccc}
\toprule
\textbf{Validation} & \textbf{Test} & \textbf{Precision} & \textbf{Recall} & \textbf{F1} & \textbf{Accuracy} \\
\midrule
2 & 8 & 0.716 & 0.724 & 0.714 & 0.724 \\
3 & 9 & 0.667 & 0.670 & 0.663 & 0.670 \\
4 & 1 & 0.746 & 0.709 & 0.722 & 0.709 \\
5 & 6 & 0.747 & 0.773 & 0.745 & 0.773 \\
6 & 4 & 0.773 & 0.795 & 0.778 & 0.795 \\
7 & 3 & 0.796 & 0.799 & 0.797 & 0.799 \\
7 & 5 & 0.800 & 0.801 & 0.794 & 0.801 \\
8 & 2 & 0.797 & 0.814 & 0.796 & 0.814 \\
9 & 7 & 0.704 & 0.697 & 0.696 & 0.697 \\
\midrule
\textbf{Mean} & -- & 0.750 & 0.754 & 0.745 & 0.754 \\
\textbf{Std}  & -- & 0.044 & 0.051 & 0.047 & 0.051 \\
\bottomrule
\end{tabular}
\end{table}

The performance of the \textit{Social Nose Contact} class varies noticeably across the considered splits, where an F1-score of 0.624 is achieved in Valid-2--Test-8, whereas the performance drops to 0.348 in Valid-7--Test-3. It is further observed that the most challenging class is \textit{Moving Away}, for which the F1-scores remain within the range of 0.073 to 0.194. This class constitutes only 4.4\% of the dataset and exhibits motion patterns that overlap with other behaviors, such as \textit{Following} and \textit{Solitary}. Therefore, the low F1-scores can be mainly attributed to the severe class imbalance and the ambiguity among closely related behaviors, rather than to a limitation of the temporal modeling itself.

\begin{table}[t]
\small
\centering
\caption{Per-class F1-scores across three representative splits in RatSI dataset}
\label{tab:beh_results}
\begin{tabular}{lccc}
\hline
\textbf{Class} & \textbf{Valid-2-Test-8} & \textbf{Valid-3-Test-9} & \textbf{Valid-7-Test-3} \\
\hline
Approaching          & 0.557 & 0.595 & 0.453 \\
Following            & 0.566 & 0.520 & 0.592 \\
Moving away          & 0.142 & 0.194 & 0.073 \\
Social Nose Contact  & 0.624 & 0.510 & 0.348 \\
Solitary             & 0.863 & 0.798 & 0.909 \\
\hline
\end{tabular}
\end{table}

\begin{figure}[H]
    \centering
    \includegraphics[width=\linewidth]{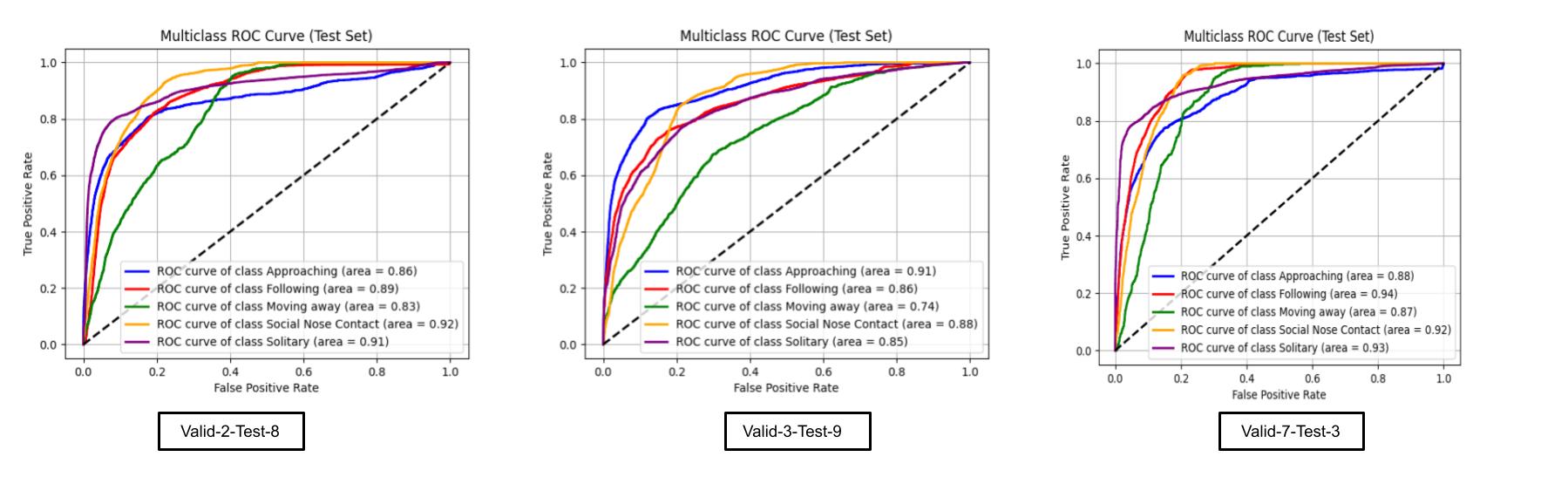}
    \caption{ROC curves across the three representative splits of the RatSI dataset.}
    \label{fig:roc_curves}
\end{figure}

The ROC curves in Figure~\ref{fig:roc_curves} provide additional insight into the model's discrimination ability. Across the three splits, AUC values range between 0.74 and 0.94. The \textit{Solitary}, \textit{Following}, and \textit{Social Nose Contact} classes achieve the highest AUC values, while \textit{Moving Away} consistently shows the lowest values. Interestingly, the AUC for \textit{Moving Away} (0.83--0.87 in two splits) is considerably higher than its F1-score, indicating that the model can rank this class correctly but struggles with the decision threshold due to severe class imbalance.

\subsubsection{Confusion Matrix Analysis}

\begin{figure}[!h]
    \centering
    \includegraphics[width=\linewidth]{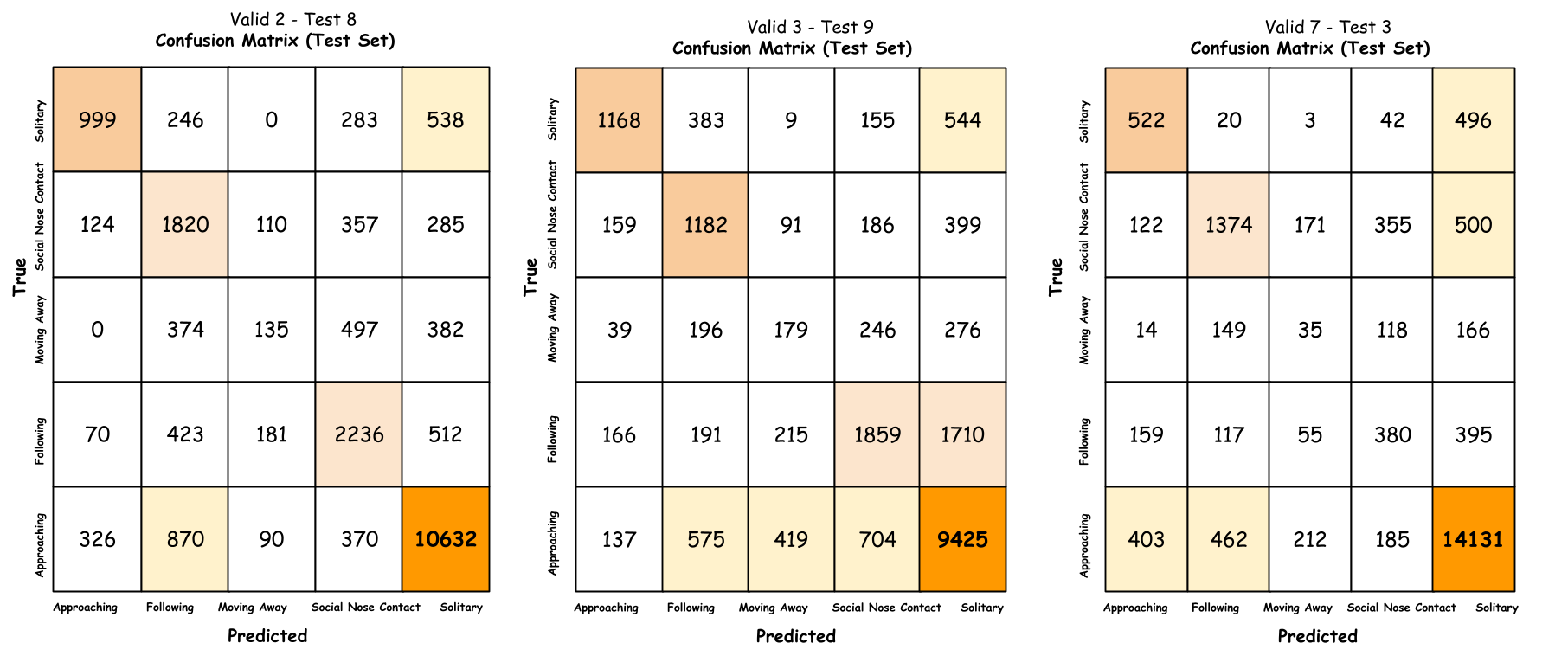}
   \caption{Confusion matrices obtained for the Valid-2--Test-8, Valid-3--Test-9, and Valid-7--Test-3 splits.}
    \label{fig:conf_matrices}
\end{figure}

The confusion matrices for the three representative splits are illustrated in Figure~\ref{fig:conf_matrices}. It is observed that, across all cases, the \textit{Solitary} class shows the most consistent performance as compared to the other classes, where more than 9,000 samples are correctly predicted in each split, and this number further increases to above 14,000 in the high-performing configuration, namely Valid-7--Test-3.

Moderate confusion is observed between \textit{Approaching} and \textit{Following}, which reflects the strong similarity in their motion patterns. It is further noted that the \textit{Social Nose Contact} class performs well in the Valid-2--Test-8 split, however, its performance degrades in the remaining splits. Moreover, the most frequent misclassification is found in the \textit{Moving Away} class, which is often predicted as \textit{Following}, \textit{Social Nose Contact}, or \textit{Solitary}. This behavior is mainly attributed to the limited number of training samples available for this class, as well as the close resemblance of its motion patterns with other  classes.

\subsubsection{Comparison with Baselines}

For baseline comparison, we selected the Valid-8--Test-2 split 
because it achieves the highest accuracy (0.814) among all splits while maintaining closely balanced precision (0.797) and F1-score (0.796), making it the most stable and representative configuration for fair baseline comparison. Using this representative configuration, we compared the proposed MSGL-Transformer with three sequential baselines: TCN~\cite{bai2018empirical}, LSTM, and Bi-LSTM~\cite{schuster1997bidirectional}. All models were trained using the same preprocessing pipeline, window length ($T=35$), optimizer, and loss function to ensure a fair comparison.

\begin{table}[ht]
\small
\centering
\caption{Comparison with different baseline models on the RatSI dataset (Valid-8--Test-2 split).}
\label{tab:model_comparison}
\begin{tabular}{lcccc}
\hline
\textbf{Model} & \textbf{Accuracy} & \textbf{Precision} & \textbf{Recall} & \textbf{F1-score} \\
\hline
TCN                   & 0.7519 & 0.7457 & 0.7519 & 0.7393 \\
LSTM                  & 0.6929 & 0.8361 & 0.6929 & 0.7392 \\
Bi-LSTM               & 0.6865 & {\bf 0.8414} & 0.6865 & 0.7377 \\
\textbf{MSGL-Transformer (Ours)} & \textbf{0.8148} & 0.7976 & \textbf{0.8148} & \textbf{0.7967} \\
\hline
\end{tabular}
\end{table}

The proposed MSGL-Transformer achieves the best overall performance, reaching an accuracy of 0.8148 and an F1-score of 0.7967. Both LSTM and Bi-LSTM perform worse, suggesting that recurrent architectures struggle to capture the complex temporal dynamics of social interactions compared with the proposed multi-scale attention mechanism.

\subsubsection{Ablation Study}

To evaluate the contribution of individual components, we conducted an ablation study on the same Valid-8--Test-2 split. Table~\ref{tab:ablation} shows that removing either the BAM module or the multi-scale attention mechanism reduces performance, confirming that both components contribute to the effectiveness of the final architecture.

\begin{table}[h!]
\small
\centering
\caption{Ablation study results on the Valid-8--Test-2 split of the RatSI dataset.}
\label{tab:ablation}
\begin{tabular}{lcccc}
\hline
\textbf{Model Variant} & \textbf{BAM} & \textbf{MSA} & \textbf{Accuracy} \\
\hline
Base  & $\times$ & $\times$ & 0.7567 \\
Base + Multi-Scale Attention (MSA) & $\times$ & \checkmark & 0.7798 \\
Base + BAM Module & \checkmark & $\times$ & 0.7859 \\
\textbf{Full Model (MSGL-Transformer)} & \checkmark & \checkmark & \textbf{0.8148} \\
\hline
\end{tabular}
\end{table}

\subsection{CalMS21 Results}

To evaluate the generalization capability of the proposed MSGL-Transformer, we further tested the model on the CalMS21 Task 1 dataset~\cite{sun2021multi}. The same architecture and hyperparameters used for RatSI were retained, with only the input dimension ($D=28$) and number of output classes ($C=4$) adapted. No other modifications were made to the model, training procedure, or loss function.

\subsubsection{Quantitative and Per-Class Results}

Table~\ref{tab:calms21_perclass} reports the per-class performance of the proposed model. The total evaluated frames (261,442) are slightly less than the raw test set (262,107) because the first $T-1 = 34$ frames of each test video cannot 
form a complete window and are excluded, along with any 
remaining incomplete windows at the end of each video. Among the four behaviors, \textit{Attack} is the most challenging, achieving an F1-score of 0.5829. This class represents only 2.8\% of the training data, making it the rarest category in the dataset.

\begin{table}[ht]
\small
\centering
\caption{Per-class results of the MSGL-Transformer on CalMS21 Task 1.}
\label{tab:calms21_perclass}
\begin{tabular}{lcccc}
\hline
\textbf{Class} & \textbf{Precision} & \textbf{Recall} & \textbf{F1-Score} & \textbf{Support} \\
\hline
Attack        & 0.5942 & 0.5720 & 0.5829 & 12,586 \\
Investigation & 0.7157 & 0.8768 & 0.7881 & 60,964 \\
Mount         & 0.8646 & 0.8798 & 0.8722 & 31,813 \\
Other         & 0.9774 & 0.8909 & 0.9321 & 156,079 \\
\hline
\end{tabular}
\end{table}

The \textit{Other} class achieves the highest F1-score (0.9321), which is consistent with its dominant representation in the dataset. The classes \textit{Mount} and \textit{Investigation} also show strong performance, with F1-scores of 0.8722 and 0.7881, respectively. In contrast, the lower performance on \textit{Attack} is likely due to a combination of two factors: its low frequency and its similarity to other close-contact behaviors, especially \textit{Investigation}. A similar trend is observed in the RatSI experiments, where the rarest class (\textit{Moving Away}) also yielded the weakest F1-scores. Together, these results indicate that rare interaction behaviors remain the main challenge across both datasets.

\begin{figure}[h]
    \centering
    \includegraphics[width=0.65\textwidth]{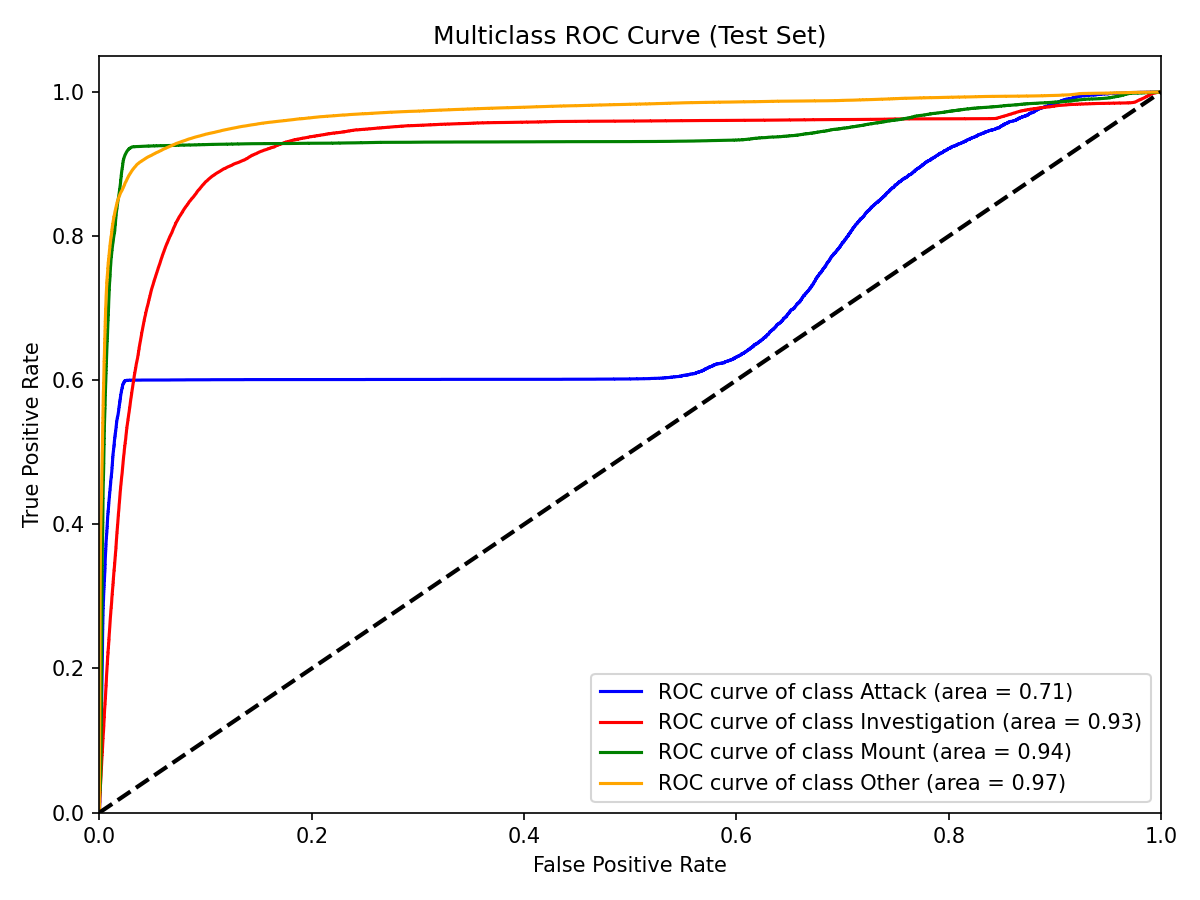}
    \caption{ROC curves of the MSGL-Transformer on the CalMS21 test set, showing per-class discrimination performance.}
    \label{fig:calms21_roc}
\end{figure}

Figure~\ref{fig:calms21_roc} shows the ROC curves for each behavior class on the CalMS21 test set. The \textit{Other} and \textit{Mount} classes achieve the highest AUC values, consistent with their strong F1-scores. The \textit{Attack} class shows the lowest AUC, reflecting the difficulty caused by its limited representation in the training data.

\subsubsection{Confusion Matrix Analysis}

\begin{figure}[h]
    \centering
    \includegraphics[width=0.72\textwidth]{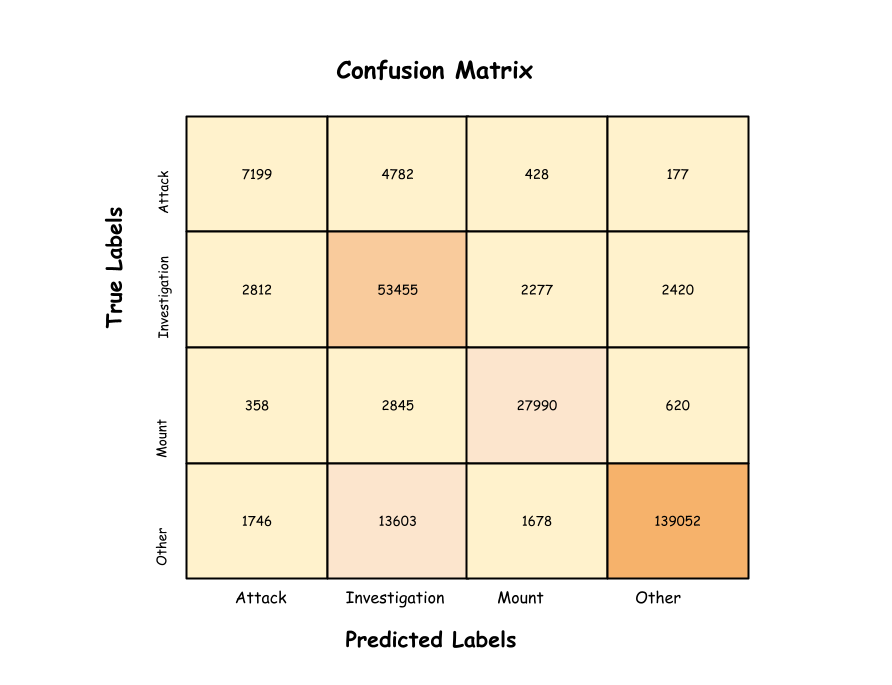}
   \caption{Confusion matrix of the MSGL-Transformer on the CalMS21 test set. We can see the model does best on the dominant \textit{Other} class, while \textit{Attack} is  the hardest class. This reflects the severe class imbalance in the dataset.}
    \label{fig:calms21_confusion}
\end{figure}

Figure~\ref{fig:calms21_confusion} shows the confusion matrix on 
the CalMS21 test set. As the prior analysis showed, the model correctly classifies the vast 
majority of \textit{Other} and \textit{Mount} frames, while 
\textit{Attack} is the most challenging class, consistent with
 its limited representation in the training set (2.8\%).

\subsubsection{Comparison with Baselines}

Table~\ref{tab:calms21_baselines} summarizes the performance of the proposed model and the baseline methods on the official CalMS21 Task 1 test set. The MSGL-Transformer achieved the highest overall accuracy (87.09\%) and F1-score (87.45\%), outperforming all baseline models. The higher performance observed on CalMS21 is likely due to the substantially larger training dataset (approximately 770k labeled frames compared with approximately 202k for RatSI), which provides stronger supervision for temporal models.

Furthermore, compared with RatSI, the performance gap between the MSGL-Transformer model and the baseline models is smaller on CalMS21, likely because the larger training set provides substantially more supervision for all methods. Nevertheless, the proposed model still achieves the best results, indicating that the multi-scale attention design remains beneficial even in a higher-data regime.

\begin{table}[ht]
\small
\centering
\caption{Baseline comparison on CalMS21 Task 1 using the official test set provided by Sun et al.~\cite{sun2021multi}.}
\label{tab:calms21_baselines}
\begin{tabular}{lcccc}
\hline
\textbf{Model} & \textbf{Accuracy} & \textbf{Precision} & \textbf{Recall} & \textbf{F1-Score} \\
\hline
TCN                   & 0.8486 & 0.8825 & 0.8486 & 0.8588 \\
LSTM                  & 0.8683 & 0.8714 & 0.8683 & 0.8696 \\
Bi-LSTM               & 0.8606 & 0.8682 & 0.8606 & 0.8638 \\
\textbf{MSGL-Transformer (Ours)} & \textbf{0.8709} & \textbf{0.8842} & \textbf{0.8709} & \textbf{0.8745} \\
\hline
\end{tabular}
\end{table}

\subsubsection{Ablation Study}

Since MSGL-Transformer was originally developed and optimized on RatSI, the ablation study on CalMS21 is better interpreted as an assessment of cross-dataset generalization rather than as a fully controlled analysis of the model components. The main reason behind this interpretation is that both datasets differ significantly in terms of scale, keypoint configuration, and class distribution. CalMS21 contains 89 videos, whereas RatSI includes only 9 videos, and, moreover, CalMS21 utilizes 28 input keypoints instead of 12. In addition, CalMS21 exhibits a more severe class imbalance, where the \textit{Other} class accounts for 62.7\% of the total data. Under these conditions, the individual effect of modules such as MSA and BAM becomes less directly visible. Even so, the full MSGL-Transformer still achieves the highest F1 score for \textit{Attack} (0.5829), even though this is the rarest behavior and accounts for only 2.76\% of the training frames. It is also important to note that MSA and BAM do not improve all metrics when they are used separately, and their main benefit appears when they work together, where BAM first modulates the features and MSA can then attend to them more effectively (Table~\ref{tab:calms21_ablation}). These results show that the combination of MSA and BAM is still beneficial for recognizing difficult and underrepresented behaviors, even in an out-of-domain setting.

\begin{table}[ht]
\small
\centering
\caption{Ablation study on CalMS21. The full model 
achieves the highest F1 for Attack, the most class-imbalanced 
behavior (2.76\% of training frames).}
\begin{tabular}{lcccc}
\toprule
\textbf{Model Variant} & \textbf{BAM} & \textbf{MSA} & \textbf{Accuracy} & \textbf{Attack F1} \\
\midrule
Base (no MSA, no BAM) & $\times$ & $\times$ & 0.8771 & 0.5635 \\
Base + MSA only       & $\times$ & \checkmark & 0.8677 & 0.5518 \\
Base + BAM only       & \checkmark & $\times$ & 0.8786 & 0.5221 \\
\textbf{Full (MSA + BAM)} & \checkmark & \checkmark & \textbf{0.8709} & \textbf{0.5829} \\
\bottomrule
\end{tabular}
\label{tab:calms21_ablation}
\end{table}

\subsubsection{Boundary Error Analysis}

To better understand the conditions under which the model fails, prediction accuracy is analyzed as a function of the temporal distance from the behavior transition points in the CalMS21 test set. A transition point is defined as the frame at which the ground-truth label changes from one behavior class to another. Across the 19 test videos, a total of 2,400 such transition events are identified.

\begin{figure}[h]
\centering
\includegraphics[width=\textwidth]{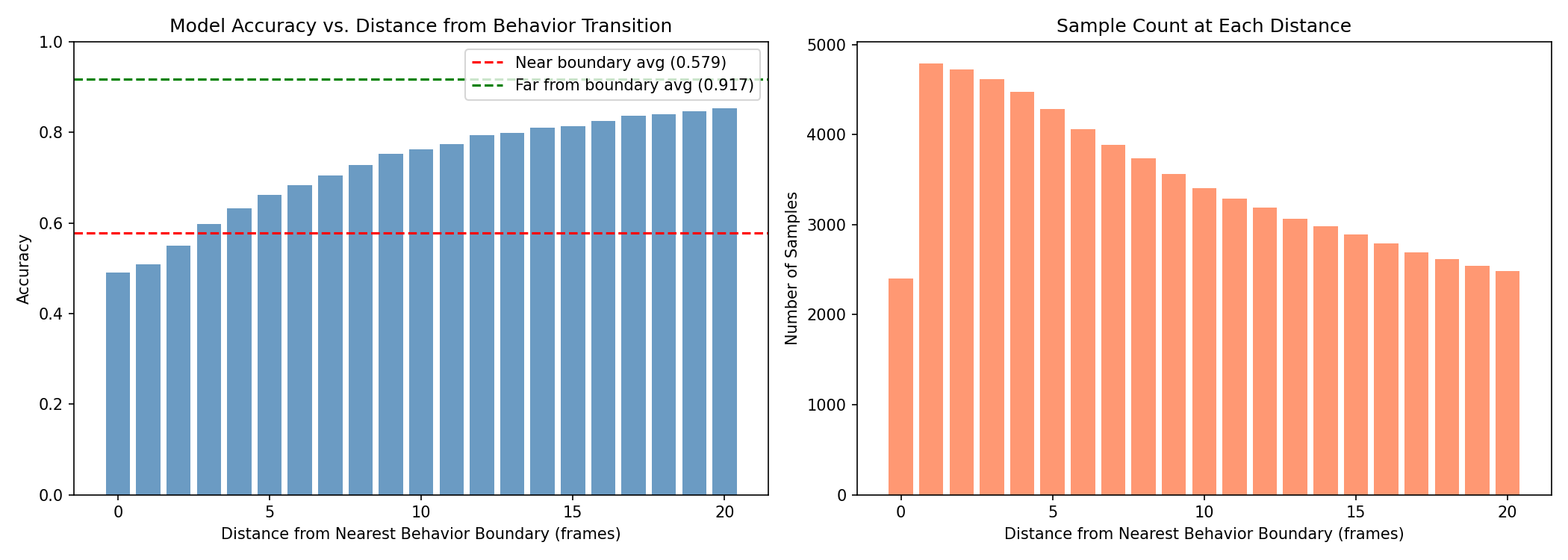}
\caption{Boundary error analysis results on the CalMS21 dataset. The left plot illustrates the variation in prediction accuracy with respect to the distance from the nearest behavior transition, while the right plot presents the distribution of samples observed at each distance.}
\label{fig:boundary}
\end{figure}

As shown in Figure~\ref{fig:boundary}, the classification accuracy is highly dependent on the proximity to the behavioral boundaries. The frames that are exactly located at the transition point are correctly classified in only 49.0\% of the cases, whereas the frames that are more than 10 frames away from a transition achieve an accuracy of 91.7\%. Moreover, the frames that lie within five frames of a transition attain an average accuracy of 57.9\%, which reflects a drop of 33.8 percentage points when compared to temporally stable segments. These results show that most of the classification errors occur near the behavioral transitions, while the proposed model performs more reliably during the stable behavioral periods.

\vspace{0.3cm}

Among the individual classes, \textit{Other} shows maximum decrease in performance near the transition boundaries. Its accuracy is reduced from 94.1\% for the frames far from transitions to 37.1\% for the frames near transitions. This behavior is expected because \textit{Other} is often used as a default label between the more specific social behaviors, therefore its boundaries are naturally ambiguous. A similar effect is also observed for \textit{Mount}, but this decrease is less severe, where the accuracy is dropped from 91.3\% to 56.9\%. 

In contrast, \textit{Attack} shows only a small decrease, from 57.4\% to 54.6\%. This suggests that the main difficulty in recognizing \textit{Attack} is not boundary ambiguity but the limited number of training samples. The most frequent transition errors occur between \textit{Other} and \textit{Investigation}, with 4,474 such errors. This finding is consistent with the per-class results, where \textit{Investigation} has the second-lowest F1-score. These two behaviors often transition into each other during natural mouse interactions, and the sliding-window formulation may include frames from both classes near the boundary.

Overall, this analysis suggests that future improvements may come from boundary-aware training strategies, such as transition-sensitive losses or explicit boundary detection modules.

\subsubsection{Comparison with Published Keypoint-Based Methods}

Table~\ref{tab:calms21_sota} presents a comparison between the proposed model and the previously reported keypoint-based methods on CalMS21 Task 1. In this regard, the results reported by Ru and Duan~\cite{ru2024hierarchical} are utilized, where several skeleton-based action recognition methods were adapted for rodent behavior recognition. Since all the compared methods use the same pose-keypoint input, this comparison is reasonably controlled.

\begin{table}[ht]
\small
\centering
\caption{Comparison with published keypoint-based methods on CalMS21 Task 1. Results for methods marked with $\dagger$ are reported from \cite{ru2024hierarchical}.}
\label{tab:calms21_sota}
\begin{tabular}{lccc}
\hline
\textbf{Method} & \textbf{Input Type} & \textbf{Avg. Per-Class Acc (\%)} & \textbf{Year} \\
\hline
ST-GCN$^\dagger$    & Keypoints & 74.5 & 2018 \\
MS-G3D$^\dagger$    & Keypoints & 73.5 & 2020 \\
CTR-GCN$^\dagger$   & Keypoints & 71.6 & 2021 \\
STGAT$^\dagger$     & Keypoints & 73.9 & 2021 \\
HSTWFormer$^\dagger$ \cite{ru2024hierarchical} & Keypoints & 76.4 & 2024 \\
\textbf{MSGL-Transformer (Ours)} & \textbf{Keypoints} & \textbf{87.1} & \textbf{N/A} \\
\hline
\end{tabular}
\end{table}

The MSGL-Transformer achieves 87.1\% average per-class accuracy, outperforming HSTWFormer by +10.7 percentage points and CTR-GCN by +15.5 percentage points. Unlike the compared graph-based methods, the proposed model does not rely on a predefined skeletal graph. Instead, it applies multi-scale temporal attention directly to the pose-coordinate sequences. These results suggest that the proposed attention-based design is highly effective for capturing behavior-relevant temporal structure in rodent interaction data, even without explicit graph convolutions.

\subsubsection{Computational Complexity}

Table~\ref{tab:complexity} presents a comparison of the computational complexity of all the evaluated models. It is observed that, with 285,892 parameters, the MSGL-Transformer is larger than the sequential baseline models, however, it still remains lightweight as compared to vision-based transformers, which generally contain millions of parameters. Although the training process requires approximately 295 seconds per epoch on an NVIDIA RTX A4000 GPU due to the inclusion of the multi-scale attention module, the full model is still able to complete the training process within nearly four hours for 50 epochs.

\begin{table}[ht]
\small
\centering
\caption{Computational complexity of  models  on CalMS21 
using an NVIDIA RTX A4000 GPU.}
\label{tab:complexity}
\begin{tabular}{lrc}
\hline
\textbf{Model} & \textbf{Parameters} & \textbf{Time/Epoch (s)} \\
\hline
TCN & 30,788 & $\sim$97 \\
LSTM & 57,604 & $\sim$87 \\
Bi-LSTM & 147,972 & $\sim$104 \\
MSGL-Transformer (Ours) & 285,892 & $\sim$295 \\
\hline
\end{tabular}
\end{table}

\subsection{Cross-Dataset Analysis}

Table~\ref{tab:cross_dataset} summarizes the performance of all evaluated models on both RatSI and CalMS21. For RatSI, we report results from the representative Valid-8--Test-2 split; for CalMS21, we report results on the official Task 1 test set.

\begin{table}[ht]
\small
\centering
\caption{Performance results of all evaluated models on the RatSI and CalMS21 datasets. For the RatSI dataset, the Valid-8--Test-2 split was used for evaluation, while for the CalMS21 dataset the Task 1 test set provided in Sun et al.~\cite{sun2021multi} was used.}
\label{tab:cross_dataset}
\begin{tabular}{lcccc}
\hline
\textbf{Model} & \textbf{RatSI Acc} & \textbf{RatSI F1} & \textbf{CalMS21 Acc} & \textbf{CalMS21 F1} \\
\hline
TCN       & 0.7519 & 0.7393 & 0.8486 & 0.8588 \\
LSTM      & 0.6929 & 0.7392 & 0.8683 & 0.8696 \\
Bi-LSTM   & 0.6865 & 0.7377 & 0.8606 & 0.8638 \\
\textbf{MSGL-Transformer (Ours)} & \textbf{0.8148} & \textbf{0.7967} & \textbf{0.8709} & \textbf{0.8745} \\
\hline
\end{tabular}
\end{table}

Across both datasets, the proposed model gives the best overall accuracy and F1-score. On RatSI, the improvement is higher because the dataset is small and the effect of multi-scale temporal modeling is more clear. On CalMS21, the difference is reduced because the larger training set helps all models for better learning. But the proposed MSGL-Transformer still gives the best overall performance.

Importantly, the architecture itself is unchanged across datasets: only the input dimensionality is adapted from 12 to 28, and the number of output classes from 5 to 4. The strong performance across two species, different pose configurations, recording conditions, and behavior vocabularies supports the claim that the model generalizes well without requiring architectural redesign.

\section{Conclusions and Future Work}

In this paper, MSGL-Transformer is presented for rodent social behavior recognition. It is a pose based sequence model. The proposed architecture combines multi-scale temporal attention with the BAM block. The purpose of this combination is to capture the complex temporal patterns present in rodent interactions. Through this proposed design, the model is able to represent rapid and subtle motion cues as well as longer interaction dynamics in a unified way. Our experimental results show that the multi-scale attention mechanism is important for modeling the layered temporal structure of rodent social behaviors, while the BAM block helps emphasize behaviorally informative temporal features. In comparison with sequential baseline models, including TCN, LSTM, and Bi-LSTM, the proposed MSGL-Transformer consistently achieved higher accuracy and F1-score values, indicating improved temporal modeling capability.

We also evaluated MSGL-Transformer on the CalMS21 dataset to assess whether the same model can generalize across a different species, pose configuration, and behavior vocabulary. The architecture and hyperparameters were kept identical to those used for RatSI, with only the input dimension and number of output classes adjusted. On CalMS21, the model achieved 87.1\% average per-class accuracy, outperforming HSTWFormer~\cite{ru2024hierarchical} by 10.7\%. It also surpassed other published keypoint-based approaches, including ST-GCN, MS-G3D, CTR-GCN, and STGAT, as reported in~\cite{ru2024hierarchical}. These results suggest that the proposed model generalizes effectively across datasets without requiring architectural redesign.

At the same time, the results also highlight an important limitation. The most challenging class in RatSI was \textit{Moving Away}, and the most challenging class in CalMS21 was \textit{Attack}; both are minority classes with substantially fewer training samples than the dominant behaviors. This indicates that class imbalance remains a key challenge for rodent behavior recognition across datasets. 

In future work, we plan to investigate class-aware loss functions and data augmentation strategies to improve recognition of minority behaviors. We also plan to evaluate the model on additional behavioral datasets to further assess its generalization ability.

\section*{Acknowledgement}
The authors acknowledge the computational resources provided by the Beocat High-Performance Computing cluster at Kansas State University (\url{https://beocat.ksu.edu}).

\section*{Funding}
This research was partially sponsored by the Cognitive and 
Neurobiological Approaches to Plasticity (CNAP) Center of Biomedical 
Research Excellence (COBRE) of the National Institutes of Health (NIH) 
under grant number P20GM113109. The content is solely the responsibility 
of the authors and does not necessarily represent the official views of NIH.

\section*{Author Contributions}
\textbf{M.I.S.}: Conceptualization, methodology, formal analysis, 
investigation, data curation, writing original draft, 
visualization. \textbf{D.C.}: Conceptualization, supervision, validation, 
writing review \& editing.

\section*{Dataset Availability}
The datasets used in this study are publicly available. The RatSI 
dataset is available at \url{https://mlorbach.gitlab.io/datasets/} 
(accessed April 2025). The CalMS21 dataset is available at 
\url{https://data.caltech.edu/records/s0vdx-0k302} 
(accessed February 2026).

\section*{Code Availability}
The code and trained models will be made publicly available on GitHub upon acceptance of this paper.

\section*{Competing Interests}
The authors declare no competing interests.

\bibliography{sample}

\end{document}